\documentclass{article}

\usepackage[numbers]{natbib}
\usepackage[final]{neurips_2024}

\usepackage[utf8]{inputenc} 
\usepackage[T1]{fontenc}    
\usepackage{hyperref}       
\usepackage{url}            
\usepackage{booktabs}       
\usepackage{amsfonts}       
\usepackage{nicefrac}       
\usepackage{xcolor}         
\usepackage{graphicx}

\title{Heuristic-Free Multi-Teacher Learning}

\author{%
  Huy Thong Nguyen\thanks{Correspondence to huythong@google.com}\\
  Google\\
  \And En-Hung Chu\\
  Google\\
  \And Lenord Melvix\\
  Google\\
  \And Jazon Jiao\\
  Google\\
  \And Chunglin Wen\\
  Google\\
  \And Benjamin Louie\\
  Google\\
}

\begin{document}

\maketitle

\begin{abstract}

We introduce Teacher2Task, a novel framework for multi-teacher learning that eliminates the need for manual aggregation heuristics. Existing multi-teacher methods typically rely on such heuristics to combine predictions from multiple teachers, often resulting in sub-optimal aggregated labels and the propagation of aggregation errors.  Teacher2Task addresses these limitations by introducing teacher-specific input tokens and reformulating the training process. Instead of relying on aggregated labels, the framework transforms the training data – consisting of ground truth labels and annotations from \textit{N} teachers – into \textit{N+1} distinct tasks: \textit{N} auxiliary tasks that predict the labeling styles of the \textit{N} individual teachers, and one primary task that focuses on the ground truth labels.  This approach, drawing upon principles from multiple learning paradigms, demonstrates strong empirical results across a range of architectures, modalities, and tasks. 

\end{abstract}

\section{Introduction}

Since AlexNet \cite{AlexNet}, a decade of ML development has yielded a wealth of capable "teachers". Humans as teachers, though expensive, provide near-perfect accuracy annotation. Large Language Models (LLMs) offer excellent zero-shot capabilities, generating high-quality "silver" data for many tasks. Domain-specific foundational models serve as specialized teachers within their domains. An ideal learning framework would enable ML models to learn effectively from all useful data sources, considering their strengths and weaknesses, unlocking the benefits of both accuracy and scalability.

However, effectively leveraging multiple teachers remains an open challenge. Conflicting annotations from humans, LLMs, and domain-specific models can be difficult to reconcile, e.g., various teachers give conflict annotation for the same input samples. Also, directly aggregating predictions from LLMs and machine learning (ML) models as final labels can be problematic due to the inherent noise in individual predictions, which can propagate and amplify in the inaccuracies after aggregation.

Existing multiple-teacher learning approaches typically leverage the aggregated output of an ensemble of teachers \cite{bookEnsemble} \cite{EnsembleMethod}. Most use a simple weighted average of teacher predictions, often with fixed or uniform weights \cite{FukudaEnsembleofTeachers}\cite{ MultiTeacherVideoRecognition}. More sophisticated approaches explore manually tuned weights \cite{EnsemblesSpeechRecognition} or learn instance-specific teacher importance weights \cite{AdaptiveMulti-TeacherMulti-levelKnowledgeDistillation} \cite{MultiTeacherBitWidth}. Alternatively, some methods focus on selecting the "best" teacher for each instance, using strategies ranging from random selection \cite{FukudaEnsembleofTeachers} to reinforcement learning-based dynamic selection \cite{ReinforcedMultiTeacherSelection}.  Specialized approaches, such as assigning teachers to distinct language pairs in multilingual neural machine translation \cite{tan2019multilingualneuralmachinetranslation}, represent specific cases of domain-based teacher selection. However, a common limitation is the reliance on pre-defined heuristics for teacher aggregation or selection, where these heuristics treat the aggregated teacher output, often noisy or sub-optimal, as the ground truth for student training. This limitation motivates our exploration of heuristic-free multi-teacher learning.

This work introduces \textbf{\textit{Teacher2Task}}, a novel multi-teacher learning method that departs from the conventional heuristic-based approaches. Our proposed method explicitly incorporates teacher-specific tokens into the input, allowing the model to internally differentiate between individual teacher labeling styles.  For each teacher, we introduce an auxiliary task: predicting the teacher's confidence score across the entire input distribution. Given \textit{N} teachers and ground truth labels, we construct \textit{N}+1 training tasks: \textit{N} auxiliary tasks focused on predicting each teacher's confidence scores, and one primary task focused on learning the ground truth. By jointly learning from both the ground truth and the diverse predictions of multiple teachers, the student model learns a more robust and nuanced understanding of the data distribution, effectively interpolating between the ground truth and diverse teacher perspectives.

The proposed approach offers several key advantages. First, it is highly label-efficient, as each teacher prediction serves as an additional training sample. Second, by explicitly encoding teacher identities within the input, the method eliminates the need for manual aggregation or selection heuristics. Finally, it mitigates the impact of potential label inaccuracies by treating teacher confidence scores as data for auxiliary tasks rather than as absolute ground truth. Experiments across various modalities and architectures demonstrate that Teacher2Task consistently benefits from the inclusion of more teachers, showcasing improved performance and robustness.

On another perspective, we extend distillation \cite{Hinton2015Distillation} from one teacher to multiple, with a relatively straightforward path to scaling to "almost infinity" teachers. Deep Learning has thrived on scaling rules, notably data-scaling and model-scaling. In a sense, we're introducing teacher-scaling, potentially opening new avenues for innovation.

\section{Proposed Heuristic-Free Multi-Teacher Learning}

\begin{figure}[t]
    \centering
    \includegraphics[width=1.0\textwidth]{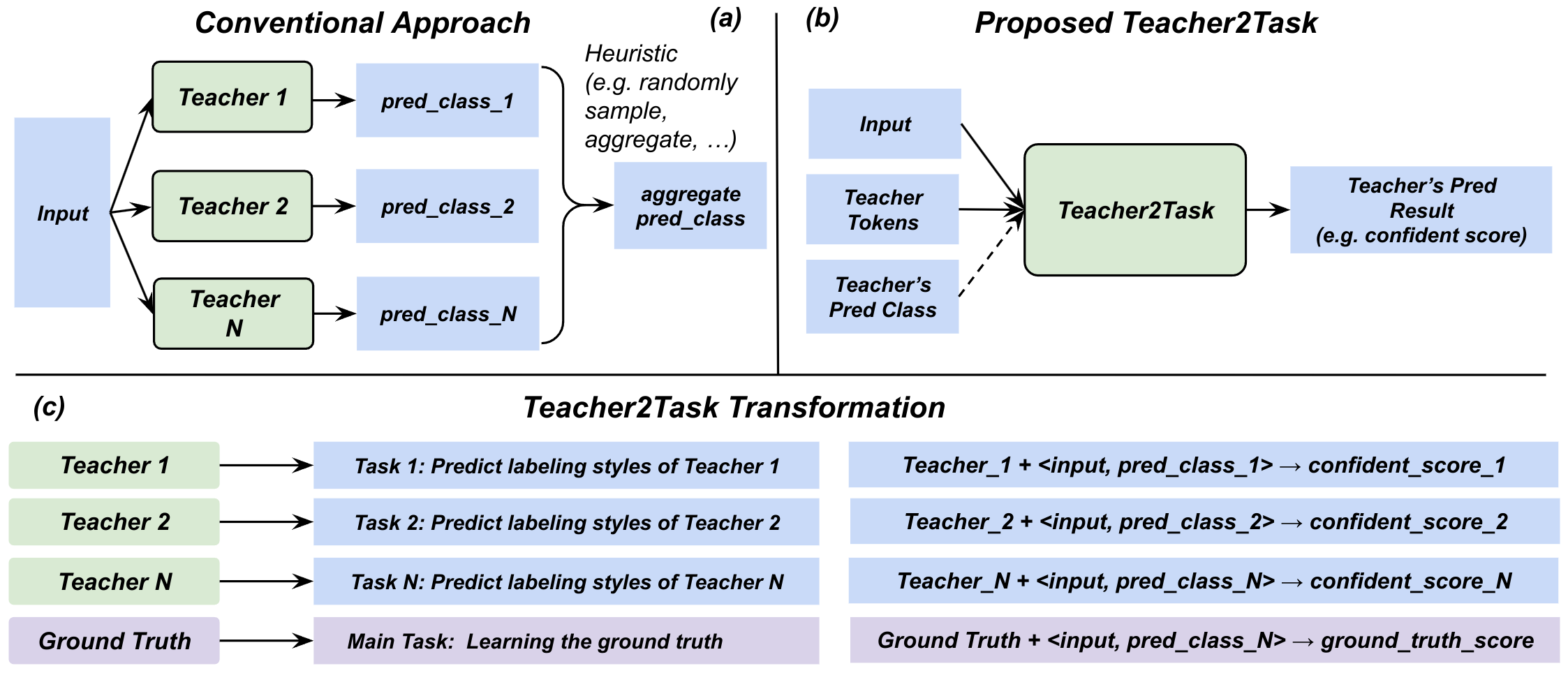}
    \caption{(a) Conventional methods with a heuristic to aggregate multiple predictions, (b) Our proposed Teacher2Task method (c) Examples of our Teacher2Task transformation.}
    \label{fig:HighLevelIdeas}
\end{figure}

\subsection{Multi-Teacher Transformation}

Traditional ensemble methods \cite{bookEnsemble}\cite{EnsembleMethod} rely on aggregating predictions from multiple teacher models for a given input (see Fig. \ref{fig:HighLevelIdeas}a). However, this approach suffers from several drawbacks:(1) \textit{Imperfect Predictions}:  Individual teacher predictions and the aggregated result can both be inaccurate. (2) \textit{Heuristic Aggregation}: Combining annotations often relies on manual and sub-optimal heuristics, and (3) \textit{Low Efficiency}:  Annotating a single sample requires running inference on all N teachers.

To address these limitations, we propose \textbf{Teacher2Task} (Fig.\ref{fig:HighLevelIdeas}b). Instead of directly aggregating predictions, we transform the problem by incorporating teacher identity and predicted class as inputs to a model that predicts the teacher's confidence score.

\begin{quote}
\textit{Multi-Teacher Input} = Teacher Identity + Original Input + Predicted Output Class
\textit{Multi-Teacher Output} = Confidence Score
\end{quote}

The proposed algorithm, though simple, introduces several key features that address inherent challenges in existing multi-teacher learning approaches.

\textbf{Individualized Teacher Tasks}: For each input sample annotated by a teacher, we add special teacher tokens to the input and train the model to predict that teacher's confidence score. This allows seamless integration of new teachers – each new teacher simply introduces a new auxiliary task:

\begin{quote}
    \textbf{Task for a new teacher}: Predict the teacher's confidence score for each input across the entire input distribution. (Fig. \ref{fig:HighLevelIdeas}c)
\end{quote}

\textbf{Resolving Annotation Conflicts}: Traditional multi-teacher learning often relies on heuristics like weighted aggregation or teacher selection to resolve conflicting annotations from multiple teachers on the same input.  Our algorithm circumvents this issue. By appending a unique teacher-specific token to each input, the model learns to differentiate between teachers and their individual labeling styles, implicitly resolving conflicts.

\textbf{Mitigating Label Noise}: Another challenge in multi-teacher learning is the potential for noisy or inaccurate labels, both from individual teachers and from aggregated predictions.  Existing methods frequently use aggregated results as pseudo-labels for the student, propagating these inaccuracies.  Our framework, however, treats teacher predictions as targets for auxiliary confidence prediction tasks. First, because all neurons in ML models are fixed, there exists an absolute mathematical formulate that transforms <inputs, output class> to <confident score>. Second, the true, human-annotated ground truth labels remain the primary learning objective. This distinction, enabled by the unique teacher tokens, allows the model to learn from both the ground truth and the diverse perspectives of multiple teachers, improving its ability to predict with human-level confidence.

\textbf{Improved Label Efficiency}: Our approach also offers significant gains in label efficiency. While aggregation methods require multiple predictions per training sample, our method generates a multi-teacher training sample from each individual teacher's prediction, reducing computational overhead.

\subsection{Conceptual Illustration}

\begin{figure}[ht]
    \centering
    \includegraphics[width=0.4\textwidth]{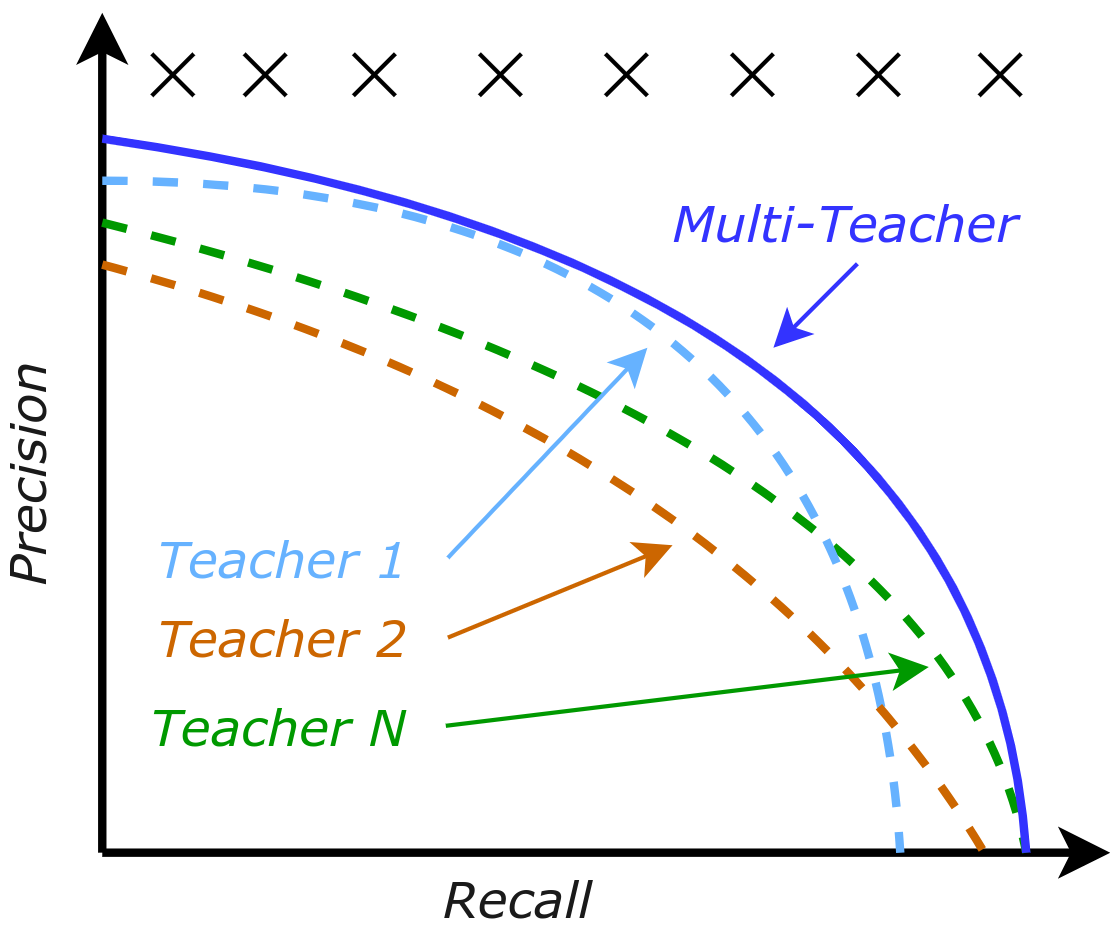}
    \caption{ Conceptual illustration for our proposed Multi-Teacher Learning. Our algorithm defines \textit{N + 1} learning tasks: \textit{N} auxiliary tasks focused on predicting each teacher's confidence scores, and one primary task focused on learning the ground truth.}
    \label{fig:ConceptualPlot}
\end{figure}

Fig.\ref{fig:ConceptualPlot} provides a conceptual illustration. Assuming a limited set of ground truth training data (black $\times$ symbols) and a need for broad generalization, we consider a scenario with N available teachers. Our algorithm defines \textit{N + 1} learning tasks: predicting the confidence scores of each teacher (N tasks for N teachers) and learning from the ground truth labels (one task).

With sufficient training data per teacher, the model learns to approximate each teacher's performance characteristics (e.g., precision-recall curves). Even a less accurate teacher (e.g., Teacher 2) provides valuable training data for its auxiliary prediction task. As Teacher2Task does not treat those less accurate labels as a part of the conventional aggregated labels, the method does not propagate inaccurate pseudo-labels for the main ground-truth learning task.

Because ground truth data is limited, the model leverages its learned understanding of the more accurate teachers, combined with the high-quality ground truth, to potentially surpass the performance of any individual teacher.  Our experimental results, showing in the \textit{Experiments} section, demonstrate this ability to interpolate towards an upper performance bound.

\subsection{Constructing Multi-Teacher Training Samples}

\begin{figure}[t]
    \centering
    \includegraphics[width=1.0\textwidth]{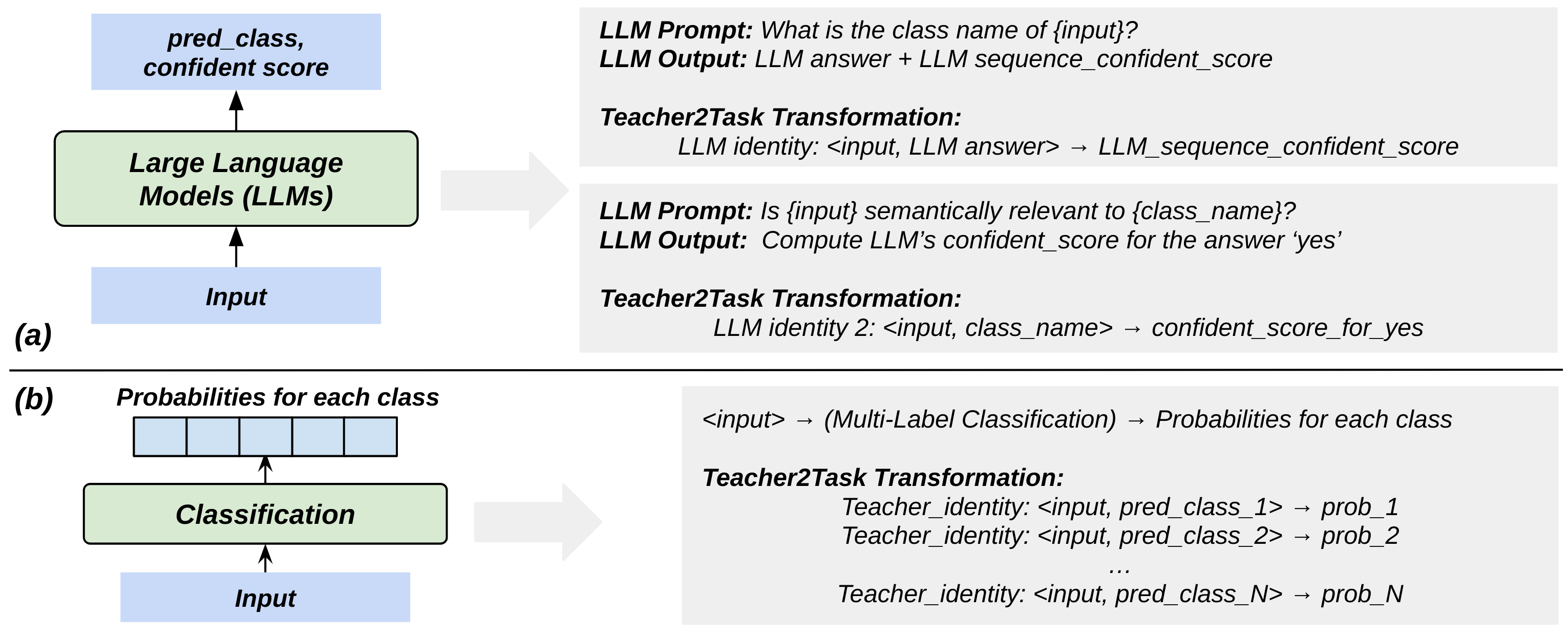}
    \caption{Examples of extracting Teacher2Task samples from (a) LLMs (b) classification models.}
    \label{fig:MultiTeacherTransformation}
\end{figure}

We demonstrate the construction of Teacher2Task training samples using Generative LLMs, multi-label classification models, and human annotators.

\textbf{Generative LLMs}:  Large language models (LLMs), prompted with specific instructions, can serve as distinct teachers in our Multi-Teacher Learning framework. Open-ended prompts like "What is the class of {input}?" yield free-text predictions with associated confidence scores. Conversely, prompts like "Is {input} semantically relevant to {class name}?" pre-define the output class, focusing on the "Yes" confidence score.  Since each unique prompt effectively creates a new "teacher" from an LLM, we prefer to predefine a fixed prompt for each LLM, consistently generating Multi-Teacher Learning samples (see Fig. \ref{fig:MultiTeacherTransformation}(a)) from each LLM inference.

\textbf{Classification Models}: For each class, we use the input, teacher name, and the model's predicted probability for that class. We can generate multiple Multi-Teacher Learning samples from a single inference of multi-class models (See Fig. \ref{fig:MultiTeacherTransformation}(b)).

\textbf{Human Annotation}: Human annotators provide relevance scores for input-class pairs. Averaging multiple annotations yields the final score.

\subsection{Model Training}

\begin{figure}[t]
    \centering
    \includegraphics[width=1.0\textwidth]{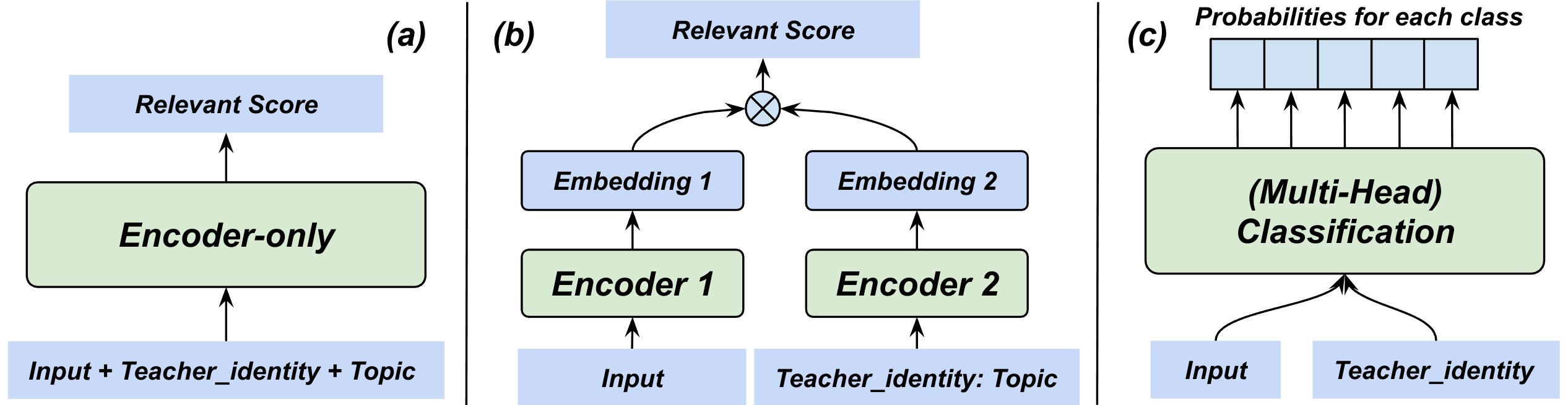}
    \caption{Various model architectures that the proposed algorithm supports (a) Encoder-only (b) Dual-Encoders (c ) (Multi-head) Classification.}
    \label{fig:ModelArchitecture}
\end{figure}

Our proposed multi-teacher learning supports various architectures (see Fig. \ref{fig:ModelArchitecture}), including: \textit{Encoder-Only} - The teacher identity can be appended directly to the input, \textit{Dual-Encoder} - Teacher information can be integrated into either the input or class topic, \textit{(Multi-Head) Classification} - Adding teacher identity to the input distinguishes label sources.

We generally utilize MSE loss on the confidence score for robust training across architectures, but Binary Cross Entropy Loss also works well. At inference, Teacher2Task allows us to predict the confidence score that any trained teacher would assign to an input-class pair. To maximize performance, we typically default to the most accurate teacher, often human annotators.

\section{Experiments}

We experiment on the open-vocabulary classification tasks for image and video understanding. Our goal is NOT to compare with public benchmarks, rather we'd like to demonstrate that our Teacher2Task algorithm effectively integrates knowledge from diverse sources to surpass that of original teachers.

\subsection{Teachers}

Our framework's flexibility allows for learning from diverse teacher types. To maximize knowledge acquisition from various label sources, we utilize four main teacher categories.

\textbf{Large Language Models (LLMs)}: We employ PaLI \cite{chen2023pali}\cite{chen2023palixscalingmultilingualvision} and Gemini \cite{geminiteam2024geminifamilyhighlycapable} and follow Section 2.3 to generate multi-teacher samples for image-topic pairs. We use slightly modified prompts for PaLI and Gemini:

\begin{quote}
PaLI: Is “{topic}” the primary focus of this image?
 
Gemini: Answer strictly with YES/NO. Does this image provide visual evidence for the topic “{topic}”?
\end{quote}

\textbf{Domain-Specific Models}: The domain-specific models excel within specific domains, providing potentially high-quality knowledge on a subset of the OpenVocab distribution.  Compared to often expensive LLMs, these models offer a cost-effective alternative to generate training samples and add knowledge on a smaller domain distribution.

\textbf{Humans}: We utilize high-quality human annotations as a key source of ground truth for our model. 

\textbf{Self-Training Teachers}: We leverage self-training \cite{xie2020selftrainingnoisystudentimproves} \cite{hieu2021MetaPseudoLabels} as a teacher type within our framework. This allows the model to iteratively learn from its own predictions on unlabeled data.

\subsection{Experiment Setup}

\textbf{Model Architectures}: We utilize the compact T5/mT5 \cite{paperT5}\cite{mt5Paper} architecture for our encoders. Images are directly converted to embedding, while videos are transformed into sequences of frame-level embedding before processing.

\textbf{Datasets}: Our human-annotated dataset consists of approximately 2M image/video samples, labeled by trained operators to indicate the presence of specific visual evidence for given topics (e.g. evening with loved ones, ancient civilizations, sriracha loaded fries, ninja warrior party, ...). We generates 200M PaLI-labeled, 20M Gemini-labeled, and 300M domain-specific ML-labeled samples.

\textbf{Adding Teacher's Identities}: Teacher2Task leverages teacher identities as input features, enabling the model to learn the distinct prediction patterns of each teacher for a given input-output pair.  For text-based models like T5/mT5 \cite{paperT5}\cite{mt5Paper}, we prepend the teacher's identifier to the input text (e.g., \textit{PaLI: {input text}} or \textit{Gemini: {input text}}). For non-textual models, such as a ResNet \cite{ResNet} for image classification, a one-hot vector representing the teacher's identity can be appended to the input.

\textbf{Evaluation}: To assess open-vocabulary generalization, we perform a topic-split evaluation to ensure that the majority of topics present in the evaluation set are unseen during training, offering a more challenging and realistic evaluation compared to conventional random splits.

\textbf{Training}: We use a learning rate of $1e^{-3}$ with a batch size of 65k throughout our experiments.

\textbf{Metrics}:  We focus on precision and recall as key performance indicators, employing PR-AUC (Precision-Recall Area Under the Curve) as our primary evaluation metric.

\subsection{Results}

This section reports experiment results of a dual-encoder configuration with a 64-dimensional embedding space, a popular choice for large-scale tagging and high-traffic retrieval systems.

\subsubsection{Image Classification}

\begin{figure}[t]
    \center
    \includegraphics[width=0.6\textwidth]{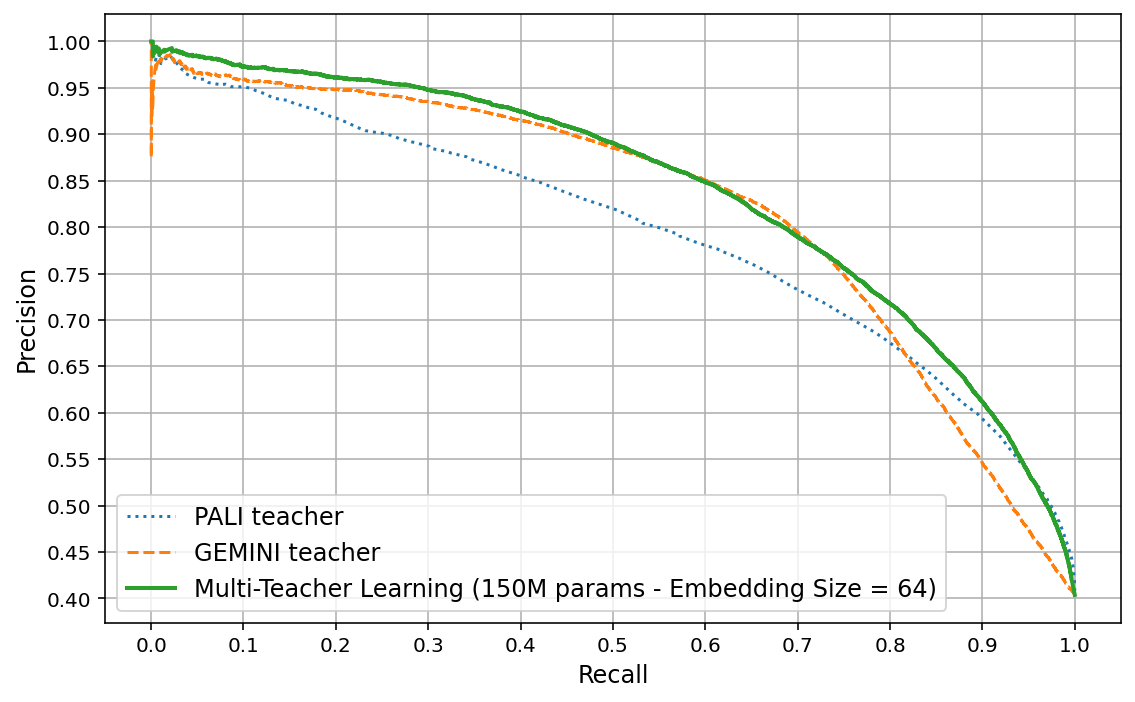}
    \caption{Precision-Recall curves comparison among PaLI, Gemini, and our Multi-Teacher Learning model. At higher precision levels, our model outperforms Gemini due to its access to human annotations. At lower precision levels, it leverages the strengths of both PaLI and Gemini, achieving an outer bound on their respective PR curves.}
    \label{fig:ImagePRCurve}
\end{figure}

Our Teacher2Task model learns image knowledge from five primary teachers:  Human annotations, PaLI \cite{chen2023palixscalingmultilingualvision}, Gemini \cite{geminiteam2024geminifamilyhighlycapable}, and 2 domain-specific models. Utilizing the specified prompts, zero-shot PaLI achieves 79.1\% PR-AUC, while Gemini reaches 82.2\% PR-AUC. Gemini, the larger model, outperforms PaLI, particularly in higher precision regions.

Our Multi-Teacher Learning model, only 150M parameters, surpasses even its best ML teacher (Gemini) by achieving 84\% PR-AUC. This highlights the benefits of combining diverse knowledge sources, even when some are imperfect.

Analyzing the PR curve reveals insightful trends (See Fig.\ref{fig:ImagePRCurve}). At higher precision levels, our model outperforms Gemini due to its access to human annotations. At lower precision levels, it leverages the strengths of both PaLI and Gemini, achieving an outer bound on their respective PR curves. This empirical results well match the conceptual illustration analyzed in Section 2.2.

\subsubsection{Video Classification}

\begin{table}[h]
\caption{Video experiment results versus the number of teachers. Clear metric wins when scaling more teachers to model training.}
\label{tableVideoMetrics}
\centering
\begin{tabular}{ccc}
\hline
Teachers                                                   & \multicolumn{1}{l}{\# Teachers} & PR-AUC \\ \hline
Baseline: 1 Video Teacher (Human)                          & 1                               & 75.6\%                           \\
1 Video Teacher (Human) + 5 Image Teachers                 & 6                               & 78.1\%                           \\
2 Video Teachers (Human + Self-Training) + 5 Image Teachers & 7                              & 80.0\%                            \\ \hline
\end{tabular}
\end{table}

Our baseline for video open-vocabulary classification, trained solely on human-annotated video-topic pairs, achieves a PR-AUC of 75.6\%.  We demonstrate the effectiveness of our Teacher2Task algorithm by scaling the number of teachers in two scenarios, both yielding metric improvements (see Table \ref{tableVideoMetrics}).

\textbf{Image-Teacher Learning}: Leveraging cross-domain knowledge transfer, we incorporate five image-based teachers (humans, LLMs and domain-specific image models), treating images as single-frame videos. Combining these with a human-annotated video teacher within our proposed algorithm increases the PR-AUC from 75.6\% to 78.1\%.

\textbf{Self-Training Integration}:  Further expanding the number of useful teachers, we integrate a self-training teacher \cite{xie2020selftrainingnoisystudentimproves} \cite{hieu2021MetaPseudoLabels} that iteratively generates pseudo-labels on unlabeled data. This expanded learning, consisting of the original human-annotated video teacher, five image teachers, and the \textit{self-training} teacher, further boosts the PR-AUC from 78.1\% to 80\%. Unlike conventional \textit{self-training}, our approach benefits from the diverse expertise of multiple teachers and explicitly distinguishes \textit{self-training} samples from ground truth labels through unique teacher identity tokens.  Also, we observe consistent performance gains with each \textit{self-training} iteration.

\subsection{Ablation Study}

Next, we run ablation study of the algorithm with various embedding sizes, model architectures, and model sizes.

\subsubsection{Embedding Sizes}

The PR-AUC increases for both image and video benchmarks when increasing the embedding size, as expected because larger embedding sizes imply better representation capabilities (see Table \ref{tableEmbeddingSize})

\begin{table}[ht]
\caption{Image \& video experiment results versus embedding sizes.}
\label{tableEmbeddingSize}
\centering
\begin{tabular}{ccc}
\hline
Embedding Size & Image PR-AUC & Video PR-AUC \\ \hline
16  & 81.2\% & 78.2\% \\
32  & 83.1\% & 79.5\% \\
64  & 84.0\% & 80.0\% \\
128 & 84.6\% & 80.3\% \\
256 & 85.2\% & 80.5\% \\ \hline
\end{tabular}
\end{table}

\subsubsection{Model Architectures}

When changing model architectures from \textit{dual-encoder} to \textit{encoder-only}, we see the encoder-only configuration slightly outperforms the dual-encoder configuration, explained by the \textit{encoder-only} can be viewed as \textit{dual-encoder} with approaching infinite embedding size (see Table \ref{tableModelArchitecture}). 

\begin{table}[h!]
\caption{Comparison of results among model architectures.}
\label{tableModelArchitecture}
\centering
\begin{tabular}{llcc}
\hline
Model Architecture & Embedding Size & Image PR-AUC & Video PR-AUC \\ \hline
Dual-Encoder       & 64             & 84.0\%              & 80.0\%                \\
Encoder-Only       & --             & 85.8\%              & 81.6\%              \\ \hline           
\end{tabular}
\end{table}

\subsubsection{Model Sizes}

Variations in model size result in minimal changes in performance. We attribute the algorithm's stable performance across model sizes to its distillation-like approach, which enables smaller student models to achieve comparable results to larger teacher models (see Table \ref{tableModelSize}).


\begin{table}[ht]
\caption{Image \& video experiment results among various model sizes.}
\label{tableModelSize}
\centering
\begin{tabular}{cccc}
\hline
Model Size & Embedding Size & Image PR-AUC & Video PR-AUC \\ \hline
150M       & 64             & 84.0\%              & 80.0\%                                  \\
300M       & 64             & 84.1\%              & 80.1\%                                  \\ \hline
150M       & 128            & 84.6\%              & 80.3\%                                  \\
300M       & 128            & 84.7\%              & 80.4\%                                  \\ \hline
\end{tabular}
\end{table}

\section{Discussion}

\subsection{Comparison to common ML algorithms}

This section positions our Teacher2Task algorithm within the broader landscape of Deep Learning methodologies, highlighting its advantages and connections to existing techniques.

\textbf{Distillation}: While distillation \cite{Hinton2015Distillation} methods typically learn from a single, often stronger, teacher, Teacher2Task aggregates knowledge from multiple sources, including human annotations and many diverse models. Our approach offers a scalable path to integrating knowledge from an "almost infinite" number of teachers.

\textbf{Ensemble Methods}:  Ensemble methods \cite{bookEnsemble}\cite{EnsembleMethod} often suffer from limitations such as reliance on manual aggregation heuristics, suboptimal aggregation strategies, and low annotation efficiency. Teacher2Task addresses these challenges by directly learning a unified model from multiple teachers within a principled framework.

\textbf{Self-Training}:  Self-training \cite{xie2020selftrainingnoisystudentimproves} \cite{hieu2021MetaPseudoLabels}, a semi-supervised technique, iteratively trains teacher-student models on labeled and pseudo-labeled data. However, it can be susceptible to confirmation bias if pseudo-labels are inaccurate \cite{arazo2020pseudolabelingconfirmationbiasdeep}. Teacher2Task mitigates this risk by separating out the source of annotations in the inputs, so that \textit{self-training} can be an additional teacher in our framework.

\textbf{Pretraining}: While Self-Supervised Learning \cite{devlin2018bert} \cite{chen2020simple} has been widely adopted for pretraining with massive unlabeled datasets, our heuristic-free multi-teacher learning offers a compelling alternative. From readily available LLMs, domain-specific models, and running those on unlabeled data, we can generate billions or even trillions of multi-teacher labeled samples for effective pretraining, maximizing knowledge transfer by enabling the pretrained model to inherit insights from all its teachers.

\subsection{Comparison to Multi-Teacher algorithms}

This section compares our proposed method to other multi-teacher approaches, dividing into three major categories: \textit{Weighted Aggregation}, \textit{Teacher Selection}, or \textit{Domain Separation} approaches.

\textbf{Weighted Aggregation}: While uniform weights for each teacher is the most common practice \cite{FukudaEnsembleofTeachers}\cite{ MultiTeacherVideoRecognition}, research has explored more sophisticated weighting approaches, such as manually tuning weights \cite{EnsemblesSpeechRecognition} or learning instance-level teacher importance weights \cite{AdaptiveMulti-TeacherMulti-levelKnowledgeDistillation}. However, even advanced weighted averaging methods suffer from drawbacks: reduced annotation efficiency (requiring multiple teacher labels per aggregated label), increased computational overhead for label aggregation, and potential imperfections in the heuristically aggregated labels.

\textbf{Teacher Selection}: \cite{FukudaEnsembleofTeachers} randomly select a teacher for each mini-batch, while \cite{ReinforcedMultiTeacherSelection} employ reinforcement learning for dynamic teacher selection. This can be considered a special case of weighted averaging, where one teacher's weight is set to 1 and the others to 0. However, this approach still suffers from increased computational overhead, potential for suboptimal teacher selection, and the inherent limitation of treating teacher predictions as ground truth.

\textbf{Domain Separation}:  \cite{tan2019multilingualneuralmachinetranslation} employ multi-teachers for multilingual neural machine translation training, where each teacher is assigned to a distinct language pair. The problem is a special case, where all teacher domains are rigidly separated  by language pairs, removing the need for teacher's aggregation or selection. In cases of potential annotation conflicts between teachers, the method might require further heuristics for domain selection.

Our approach addresses the challenges of existing multi-teacher methods. By treating each teacher's prediction as a training sample, we maximize annotation efficiency by leveraging all teacher labels. Furthermore, incorporating teacher identities as input features and re-framing the task as predicting individual teacher labeling styles, we remove the need for weight aggregation, teacher selection, or domain separation. By not viewing teacher confidence scores as ground-truth labels, our algorithm eliminates the problem of imperfect aggregation heuristics.



\section{Conclusion}

Teacher2Task offers a unified and scalable approach that leverages the strengths of multiple learning paradigms. By utilizing confidence scores from a potentially vast number of teachers, it extends the concept of distillation while eliminating the need for explicit aggregation heuristics. The approach facilitates the generation of massive training datasets from unlabeled data, proving particularly effective for training compact, yet highly knowledgeable, student models that inherit the collective knowledge of all original teachers.

\bibliographystyle{ieee}
\bibliography{main}

\end{document}